\documentclass[10pt,twocolumn,letterpaper]{article}

\usepackage[pagenumbers]{cvpr} 
\usepackage{graphicx}
\usepackage{amsmath}
\usepackage{amssymb}
\usepackage{booktabs}
\usepackage{makecell}
\usepackage{array}
\usepackage{wrapfig}
\usepackage{multirow}
\usepackage{bbm}
\usepackage{float}
\usepackage{graphicx}
\usepackage{perpage}
\usepackage{enumitem}
\usepackage{algpseudocode}
\usepackage{extarrows}
\usepackage{amssymb}
\usepackage{booktabs}
\usepackage{marvosym}
\usepackage{amsfonts}
\usepackage{gensymb}
\usepackage{marvosym}
\usepackage{pifont}
\usepackage[table]{xcolor}
\usepackage{algorithm}
\usepackage{newfloat}
\usepackage{listings}
\floatstyle{ruled}
\newfloat{listing}{tb}{lst}{}
\floatname{listing}{Listing}

\newcommand{\yes}{\ding{51}}
\newcommand{\no}{\ding{55}}

\newcommand{\green}[1]{\textcolor[RGB]{96,177,87}{#1}}
\newcommand{\red}[1]{\textcolor[RGB]{227,96,87}{#1}}
\usepackage[pagebackref,breaklinks,colorlinks]{hyperref}

\usepackage[capitalize]{cleveref}
\crefname{section}{Sec.}{Secs.}
\Crefname{section}{Section}{Sections}
\Crefname{table}{Table}{Tables}
\crefname{table}{Tab.}{Tabs.}

\def\cvprPaperID{*****} 
\def\confName{CVPR}
\def\confYear{2022}

\begin{document}

\title{Champion Solution for the WSDM2023 Toloka VQA Challenge}
\author{
    Shengyi Gao$^{1}$,
    Zhe Chen$^{1,2}$,
    Guo Chen$^{1,2}$,
    Wenhai Wang$^{2}$,
    Tong Lu$^{1}$\textsuperscript{\Letter}\\
    $^1$Nanjing University~~~
    $^2$Shanghai AI Laboratory\\
    {\small \url{https://github.com/czczup/ViT-Adapter/tree/main/wsdm2023}}
}
\maketitle

\begin{abstract}
In this report, we present our champion solution to the WSDM2023 Toloka Visual Question Answering (VQA) Challenge.
Different from the common VQA and visual grounding (VG) tasks, this challenge involves a more complex scenario, i.e.~inferring and locating the object implicitly specified by the given interrogative question.
For this task, we leverage ViT-Adapter, a pre-training-free adapter network, to adapt multi-modal pre-trained Uni-Perceiver for better cross-modal localization.
Our method ranks first on the leaderboard, achieving 77.5 and 76.347 IoU on public and private test sets, respectively.
It shows that ViT-Adapter is also an effective paradigm for adapting the unified perception model to vision-language downstream tasks. 
Code and models will be released.
\end{abstract}

\vspace{0.5em}
\section{Introduction}
\label{sec:intro}

\begin{figure}[t]
  \centering
   \includegraphics[width=0.99\linewidth]{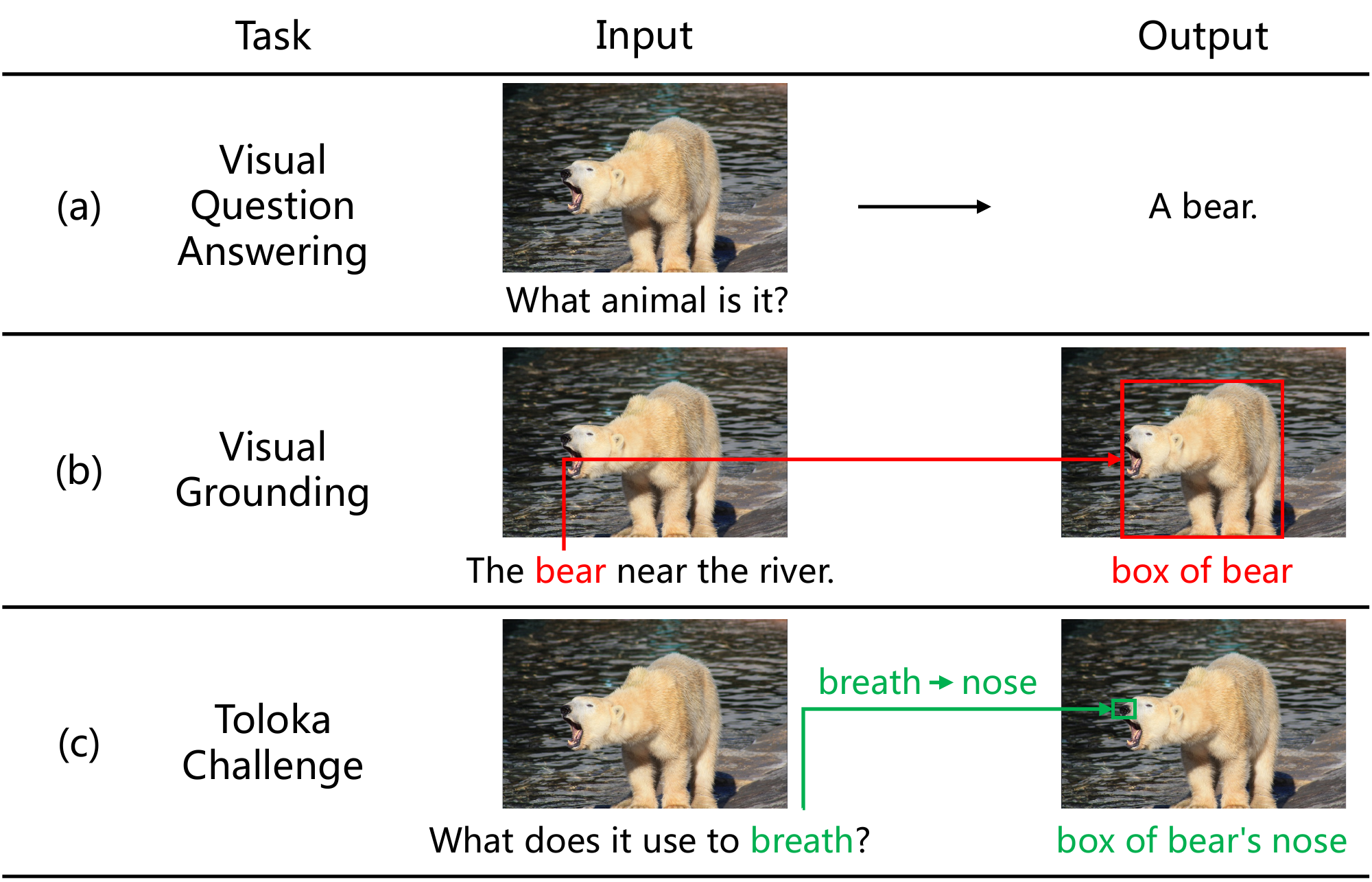}
   \caption{\textbf{Task Comparison.} Different from standard VQA and VG tasks, the competition dataset challenges us to infer and locate the object implicitly specified by the given interrogative question.}
   \label{fig:task}
\end{figure}

Visual question answering (VQA)\cite{antol2015vqa} is a well-studied task in the multi-modal field, which involves generating natural language answers according to the given images and associated questions.
Another typical task is visual grounding (VG)\cite{deng2018visual,fukui2016multimodal}, aiming to locate the target object in an image in the light of a description, usually a declarative sentence, which is related to object detection.
Different from these standard tasks in the research community,
the WSDM2023 Toloka Visual Question Answering Challenge explores a more complex task in between, \ie inferring and locating the object implicitly specified by the given \emph{interrogative} question, as shown in Figure~\ref{fig:task}(c).
In other words, the category name of objects doesn't explicitly appear in questions, which makes it more challenging than common VG tasks (\eg, RefCOCO\cite{yu2016modeling}).

In the VQA and VG tasks, the widely-used pipeline is to build a two-tower model with individual image/text backbones, along with a tailored head.
For example, ResNet \cite{he2016deep} and BERT~\cite{devlin2018bert} are often employed as the image backbone and the text backbone, respectively.
For the task-specified head, Liu \etal \cite{liu2022dq} recently designed a DETR-like head for VG tasks, namely DQ-DETR, which shows leading performance on several popular datasets.

Different from the existing pipeline, we explore a new paradigm to deal with the vision-language task in this challenge: (1) We adopt the unified model Uni-Perceiver~\cite{zhu2022uni} as the backbone, instead of individual image backbone and text backbone. 
(2) We employ ViT-Adapter~\cite{chen2022vitadapter} to arm the Uni-Perceiver, to enhance its localization capacity by introducing image-related inductive biases.
(3) We use the general object detector DINO~\cite{zhang2022dino} as the head, rather than a specifically designed head for VQA and VG tasks.

Extensive experiments on the competition dataset show the effectiveness of our solution. 
It demonstrates that in addition to uni-modal tasks, ViT-Adapter~\cite{chen2022vitadapter} is also an effective paradigm for adapting the unified perception model (\eg, Uni-Perceiver~\cite{zhu2022uni}) to vision-language downstream tasks.
As a result, our method ranks first on the leaderboard, producing 77.5 and 76.347 IoU on public and private test sets, respectively.
In the remainder of this technical report, we will introduce the detailed architecture of our solution, show the experiments and analysis, and make a summary of the experience of participating in this challenge.

\section{Related Work}
\label{sec:related}

\subsection{Visual Grounding}

Visual grounding (VG) \cite{fukui2016multimodal} is a widely researched field, which aims to locate objects conditioned on the provided textual description.
Existing methods for VG tasks can be roughly divided into two categories: two-stage and one-stage methods. Two-stage methods \cite{deng2018visual,hong2019learning,chen2021ref} first generate region proposals and then select the best matching region as results, while one-stage methods \cite{yang2019fast,du2022visual,deng2021transvg} predict the bounding boxes directly. 
For example, MDETR \cite{kamath2021mdetr} firstly attempt to improve DETR-series detectors \cite{carion2020end,zhu2020deformable,zhang2022dino} for end-to-end visual grounding.
Recently, DQ-DETR \cite{liu2022dq} designed a novel duel-query decoder framework and improves the contrastive loss in MDETR, achieving state-of-the-art performance on several popular benchmarks.
Unlike the two-stage baseline~\cite{toloka2022baseline} provided by the challenge organizer, our solution is a one-stage approach.

\subsection{Object Detection}
Object detection has been a research focus in the computer vision community.
In the past few years, CNN-based detectors dominated in this field, such as Faster R-CNN \cite{ren2015faster}, Mask R-CNN \cite{he2017mask}, HTC++ \cite{liu2021swin}, \etc.
Recently, transformer-based methods have shown incredible potential for object detection.
DETR \cite{carion2020end} is the first transformer-based end-to-end object detector, which brings a new trend of designing transformer-based detectors, but still suffers from slow convergence.
To remedy this issue, Deformable DETR \cite{zhu2020deformable} designed the deformable attention that only attends to certain sampling points around a reference point.
DINO \cite{zhang2022dino} is the current state-of-the-art DETR-like detector, which significantly improves both training efficiency and detection performance by introducing contrastive denoising training.
Nowadays, DINO has been widely-used in the detection community, such as many benchmarks~\cite{lin2014microsoft,gupta2019lvis} and competitions~\cite{chen2022internvideo,zhang20221st}. 
Inspired by this, we also adopted DINO as the detector in our solution.

\subsection{Unified Architecture}

Network architecture design is an ongoing research topic both in upstream~\cite{liu2021swin,wang2021pyramid,wang2022pvt,wang2022internimage} and downstream tasks~\cite{chen2021fast,chen2022towards,chen2020siameseccr,chen2022dcan,zheng2020dynamic}.
For vision-language tasks, the network architecture is usually designed as a two-tower paradigm~\cite{kamath2021mdetr,liu2022dq}. 
In other words, individual image/text backbones are employed to extract visual/textual features, such as ResNet \cite{he2016deep} and BERT~\cite{devlin2018bert}, respectively.
Recently, unified architecture for multi-modal tasks has been attracting increasing attention.
For instance, Uni-Perceiver~\cite{zhu2022uni} is a ViT-based unified perception architecture that can process a variety of modalities and tasks with a single model and shared parameters.
With extensive pre-training on multi-modal tasks, Uni-Perceiver shows its powerful transfer ability on various downstream tasks.
In addition, ViT-Adapter~\cite{chen2022vitadapter} is a pre-training-free adapter network, which can adapt pre-trained unified models to specific downstream tasks and achieve better performance.
Inspired by them, our solution adopts both Uni-Perceiver and ViT-Adapter, building a unified backbone for both image and text inputs.

\begin{figure}
  \centering
  \includegraphics[width=0.98\linewidth]{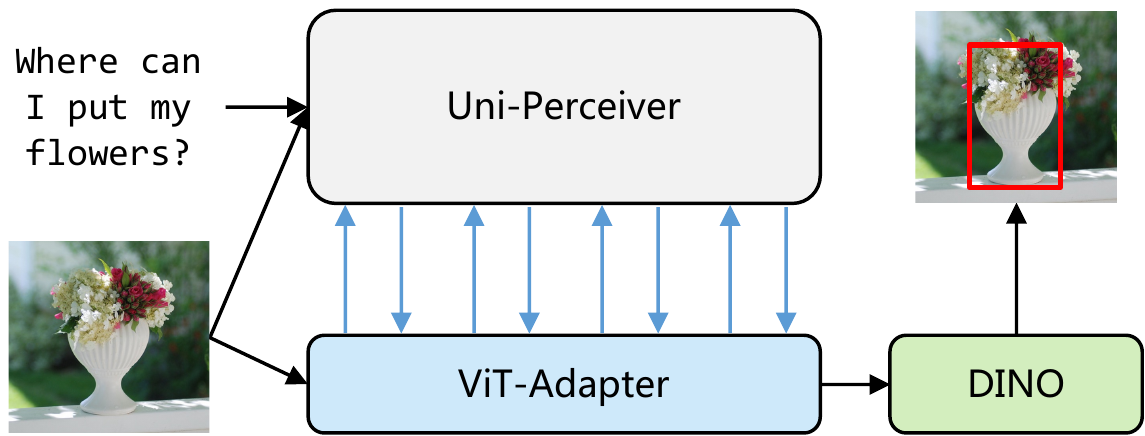}
  \caption{\textbf{Overall Architecture.} Our solution consists of three main components, including Uni-Perceiver \cite{zhu2022uni}, ViT-Adapter \cite{chen2022vitadapter}, and DINO \cite{zhang2022dino}.
  It shows that in addition to uni-modal tasks, ViT-Adapter is also an effective paradigm to adapt the unified ViT model to vision-language downstream tasks.
  }
  \label{fig:arch}
\end{figure}

\section{Methodology}
\label{sec:meth}

\subsection{Overall Architecture}
Figure~\ref{fig:arch} illustrates the overall architecture of our model, which contains three main components: Uni-Perceiver \cite{zhu2022uni}, ViT-Adapter \cite{chen2022vitadapter}, and DINO \cite{zhang2022dino}.
Among them, Uni-Perceiver is a ViT-based unified perception model, which can process a variety of modalities, such as text and image. 
ViT-Adapter utilizes the task prior and the input image, to enhance the localization capacity of the Uni-Perceiver.
Thanks to the full interaction of text and image information within the backbone, we can cast this task as common object detection and use a general detection head.

\subsection{Backbone Network}

The backbone in our solution consists of Uni-Perceiver \cite{zhu2022uni} and ViT-Adapter \cite{chen2022vitadapter}, as shown in Figure~\ref{fig:arch}.
Uni-Perceiver is a unified architecture that can handle various modalities.  
Given the raw inputs from text and image modalities, modality-specific tokenizers are applied to generate the input token sequences.
Specifically, Uni-Perceiver uses the BPE tokenizer~\cite{sennrich2016neural} and image patch tokenizer~\cite{dosovitskiy2020image} for text and image modality, respectively. 
Finally, these output tokens are concatenated into token sequences and then fed into a modality-agnostic transformer encoder.

Besides, we equip the Uni-Perceiver with ViT-Adapter, which is designed as a powerful task-specific adapter for plain ViT. It consists of:
(1) a spatial prior module to capture the local semantics (spatial prior) from the input image, (2)
a spatial feature injector to inject spatial priors into the ViT, and (3) a multi-scale
feature extractor to reconstruct hierarchical features from the ViT.
As shown in Figure~\ref{fig:arch}, ViT-Adapter conducts multiple feature interactions with Uni-Perceiver, which enhances the localization capability of the unified model.

\subsection{Detection Head}

Since the text information and image information have been fully mixed and interacted in the backbone, we can simply cast this task as single-class object detection and use a general detection head.
Given that DINO~\cite{zhang2022dino} is the current state-of-the-art detection head, and has been widely-used in many benchmarks~\cite{lin2014microsoft,gupta2019lvis} and competitions~\cite{chen2022internvideo,zhang20221st}, we also adopted DINO as the detector in our solution.

\subsection{Auxiliary Loss}
\label{subsec:loss}

In addition to the original loss in DINO~\cite{zhang2022dino}, we adopt a segmentation-based auxiliary loss for better training. 
Specifically, we employ the Semantic FPN head \cite{kirillov2019panoptic} on the features of 1/8, 1/16, and 1/32 scales to segment the region containing the specified object. 
For a given image, we use the ground truth box to generate the binary mask, and adopt Dice loss~\cite{jadon2020survey} for supervision.
It can be written as:
\begin{equation}
  \mathcal{L}=\mathcal{L}_{\rm o}+\lambda\cdot \mathcal{L}_{\rm aux},
  \label{eq:loss}
\end{equation}
where $\mathcal{L}_{\rm o}$ represents the original loss of DINO, and $\mathcal{L}_{\rm aux}$ denotes the proposed auxiliary loss. $\lambda$ is the coefficient of the auxiliary loss, which is set to 1.0 by default.

\subsection{Test-Time Augmentation}

\begin{algorithm}[!t]
\caption{PyTorch-like Pseudo Code of TTA}
\label{alg:pseudo_code}
\definecolor{codeblue}{rgb}{0.25,0.5,0.5}
\definecolor{codekw}{rgb}{0.85, 0.18, 0.50}
\lstset{
	backgroundcolor=\color{white},
	basicstyle=\fontsize{7.5pt}{7.5pt}\ttfamily\selectfont,
	columns=fullflexible,
	breaklines=true,
	captionpos=b,
	commentstyle=\fontsize{7.5pt}{7.5pt}\color{codeblue},
	keywordstyle=\fontsize{7.5pt}{7.5pt}\color{codekw},
}

\begin{lstlisting}[language=python]

def tta_inference(imgs, metas, model):
    boxes, scores = [], []
    
    for img, meta in zip(imgs, metas):
        top1_box, top1_score = model(img, meta)
        boxes.append(top1_box) # [1, 4]
        scores.append(top1_score) # [1]
        
    boxes = torch.cat(boxes, dim=0) # [n, 4]
    scores = torch.cat(scores, dim=0) # [n]

    iou_scores = iou(boxes, boxes).mean(dim=1) # [n]
    corrected_scores = scores + iou_scores # [n]

    max_index = torch.argmax(corrected_scores)
    final_box = boxes[max_index]
    final_score = scores[max_index]
    
    return final_box, final_score 

\end{lstlisting}
\end{algorithm}

For better performance, we specially design a test-time augmentation (TTA) algorithm, as shown in Algorithm~\ref{alg:pseudo_code}.
Specifically, for each sample, we adopt multi-scale test and horizontal flip during inference.
After that, we obtain $n$ different top-1 boxes and scores.
In order to accurately select the most suitable one out of $n$ boxes, 
we use these boxes to calculate the mean Intersection over Union (IoU) scores against each other, and employ it as a correction for original classification scores.
At last, the box with the highest corrected score will be selected as the final result.

\section{Experiments}
\label{sec:exp}

\subsection{Datasets}

\textbf{Toloka VQA dataset} \cite{toloka2022baseline} consists of images associated with textual questions. 
It has 45,199 instances split among three subsets: train (38,990 instances), public test (1,705 instances), and private test (4,504 instances).
One instance in the dataset is a question-image pair labeled with the ground truth coordinates of a bounding box containing the visual answer to the given question. 
And the predicted results are evaluated by the IoU metric. 
Notably, different from the widely-used visual grounding dataset RefCOCO \cite{yu2016modeling}, all texts in the Toloka VQA dataset are interrogative questions rather than declarative sentences.
In other words, the category name of objects doesn't explicitly appear in questions.

\textbf{GQA dataset} \cite{hudson2019gqa} is originally designed for VQA tasks. 
It develops a question engine to leverage Visual Genome scene graph structures to create 22M diverse reasoning questions. 
In this challenge, we construct a visual grounding dataset based on GQA for pre-training. Because of the scene graphs, it is easy to obtain the target object for each question, and we can simply construct abundant image-question pairs with bounding boxes. 
Why we choose GQA rather than RefCOCO \cite{yu2016modeling} for pre-training is that GQA offers texts in the form of interrogative questions, which is semantically closer to the Toloka VQA dataset. 
Since the GQA dataset is quite large, we filter out questions correlated to only one object in the scene graph. The final constructed dataset contains 57,519 images and 163,917 questions, split into a training set with 160,000 samples and a validation set with 3,917 samples.

\subsection{Implementation Detail}

We implement two models with different sizes, based on Uni-Perceiver-B and Uni-Perceiver-L, respectively. 

\textbf{Pre-train.} To achieve higher performance, we first pre-train our models on the GQA dataset~\cite{hudson2019gqa}. 
We initialize the backbone with pre-trained weights provided by Uni-Perceiver~\cite{zhu2022uni}.
The training image is resized to have a shorter side of 480-800 pixels, while the longer side does not exceed 1,333 pixels.
The model is trained for 6 epochs with a batch size of 16 and a constant learning rate of $1\times 10^{-4}$.
More training recipes are listed in Table~\ref{tab:para}.

\textbf{Fine-tune.} We load the GQA pre-trained weights and fine-tune our model on the Toloka VQA dataset for a 2$\times$ (24 epochs) schedule. Besides, we develop a new data augmentation technique for fine-tuning, \ie~using a pre-trained T5 model~\cite{raffel2020exploring} to paraphrase the given question with a probability of 50\%. Other training recipes for fine-tuning are kept the same as the pre-training stage, as shown in Table~\ref{tab:para}.
During testing, the shorter side of input images is fixed to 800 pixels, while the longer side does not exceed 1,333 pixels unless TTA is specifically stated.

\begin{table}[!t]
  \centering
  \small
  \setlength{\tabcolsep}{1.8mm}
  \begin{tabular}{@{}lcc@{}}
    \toprule
    \textbf{Settings} & \textbf{Base} & \textbf{Large}\\
    \midrule
    input resolution & 480-800/1333 & 480-800/1333 \\
    batch size & 16 & 16 \\
    optimizer & AdamW & AdamW\\
    epoch (pre-train/fine-tune) & 6/24 & 6/24 \\
    learning rate & $1\times10^{-4}$ & $1\times10^{-4}$ \\
    layer-wise decay rate & 0.65 & 0.8\\
    drop path rate & 0.2 & 0.3\\
    weight decay & 0.05 & 0.05\\
    \midrule
    horizontal flip  &  \yes &  \yes  \\
    auto augment     &  \yes &  \yes  \\
    paraphrase (pre-train/fine-tune)  &  \no/\yes &  \no/\yes \\
    exponential moving average   &  \yes &  \yes  \\
    \bottomrule
  \end{tabular}
  \caption{\textbf{Training Recipes.} 
  We train two models with different parameter scales, based on Uni-Perceiver-B and -L, respectively. }
  \label{tab:para}
\end{table}

\subsection{Ablation Study}

\begin{table}[!t]
  \centering
  \small
  \setlength{\tabcolsep}{1.3mm}
  \begin{tabular}{lll}
    \toprule
    \# & \textbf{Method} & \textbf{IoU (gain)} \\
    \midrule
    1 & Baseline: Uni-Perceiver-B + ViTDet (4 global) & 69.2 \\
    2 & Substitute ViTDet with ViT-Adapter & 71.1 \green{(+1.9)} \\
    3 & Replace window attention with global attention  & 72.1 \green{(+1.0)} \\
    4 & Use paraphrase augmentation  & 72.5 \green{(+0.4)} \\
    5 & Use auxiliary loss  & 72.8 \green{(+0.3)} \\
    6 & Replace Uni-Perceiver-B with Uni-Perceiver-L  & 75.7 \green{(+2.9)} \\
    7 & Use RefCOCO pre-training  & 75.2 \red{(-0.5)} \\
    8 & Use GQA pre-training & 76.6 \green{(+0.9)} \\
    \rowcolor{gray!15}
    9 & Apply our TTA algorithm  & \textbf{77.5 \green{(+0.9)}} \\
    \bottomrule
  \end{tabular}
  \caption{\textbf{Ablation Studies.} 
  We report the IoU performance of our different models on the public test set. }
  \vspace{-0.5em}
  \label{tab:abla}
\end{table}

\textbf{Effect of ViT-Adapter.} 
As reported in $\#1$ in Table~\ref{tab:abla}, in our initial experiments, we build a simple backbone based on Uni-Perceiver-B~\cite{zhu2022uni} and ViTDet~\cite{li2022exploring}.
But it still has a significant gap with other participants on the leaderboard.
Therefore, we substitute ViTDet with ViT-Adapter~\cite{chen2022vitadapter} to introduce more image-related inductive biases, obtaining +1.9 IoU improvements, as shown in $\#2$ in Table~\ref{tab:abla}.

\textbf{Effect of more global attention.}
In the beginning, we only adopt 4 global attention in Uni-Perceiver following ViTDet~\cite{li2022exploring}. However, comparing $\#2$ and $\#3$ in Table~\ref{tab:abla}, we found that using global attention in all layers outperforms 4 global attention by a large margin of +1.0 IoU.
In other words, global attention plays a more important role in the visual grounding task than common object detection.

\textbf{Effect of paraphrase augmentation.} 
In the fine-tuning stage, we use a pre-trained T5 model~\cite{raffel2020exploring} to generate synonymous texts for each question in the Toloka VQA dataset.
During training, we randomly select one of all generated questions as the text input with a probability of 50\%. 
As shown in $\#4$ in Table~\ref{tab:abla}, this technique improves the IoU performance of our model from 72.1 to 72.5.

\textbf{Effect of auxiliary loss.}
As described in Section~\ref{subsec:loss}, we proposed a segmentation-based auxiliary loss for training. 
Comparing $\#4$ and $\#5$ in Table~\ref{tab:abla}, the proposed auxiliary loss can slightly improve the performance by +0.3 IoU.

\textbf{Effect of larger model.}
We trained a larger model by replacing Uni-Perceiver-B with Uni-Perceiver-L. 
As reported in $\#6$ in Table~\ref{tab:abla}, Uni-Perceiver-L leads to a significant performance improvement of +2.9 IoU (75.7 \emph{vs.}~72.8).

\textbf{Effect of GQA pre-training.} 
We tried to pre-train our model on the RefCOCO \cite{yu2016modeling} and GQA datasets \cite{hudson2019gqa}, respectively. 
As shown in $\#7$ and $\#8$ in Table~\ref{tab:abla}, the pre-training on GQA boosts the performance of our model (76.6 \emph{vs.}~75.7), while the pre-training on RefCOCO makes a negative effect (75.2 \emph{vs.}~75.7). 
We argue that the main reason is the semantic inconsistencies between RefCOCO and Toloka VQA, since the texts in RefCOCO are all declarative, while the texts in the Toloka VQA are interrogative.

\textbf{Effect of TTA.} 
We apply the multi-scale test of three scales $\{(600, 1333),$ $(800, 1333),$ $(1000, 1333)\}$, and horizontal flip for each sample.
In other words, for each image-text pair, we obtain 6 different top-1 boxes and scores, and then employ the proposed TTA algorithm to select the most appropriate box as the final result.
As listed in $\#8$ and $\#9$ in Table~\ref{tab:abla}, our TTA algorithm further boosts the performance of our model to 77.5 IoU on the public test set.

\subsection{Comparison with Other Teams}

Table~\ref{tab:res} shows the performance of our solution compared with other teams. It can be seen that our method finally achieved 76.347 IoU on the private test set.

\begin{table}[!t]
  \centering
  \small
  \linespread{0.3}
  \setlength{\tabcolsep}{3.95mm}
  \begin{tabular}{clc}
    \toprule
    \textbf{Rank} & \textbf{Teams} & \textbf{Private Test} \\
    \midrule
    \rowcolor{gray!15}
    1 & NJU IMAGINE LAB (ours) & \textbf{76.347} \\
    2 & jinx, Zhouyang\_Chi & 76.342 \\
    3 & komleva.ep  & 75.591 \\
    4 & xexanoth  &  74.667 \\
    5 & Man\_of\_the\_year  & 72.768\\
    6 & Haoyu\_Zhang, KhylonWong  & 71.998\\
    7 &	nika-li & 70.525 \\
    8 & blinoff	& 62.037 \\
    9 & Ndhuynh	& 61.247 \\
    \bottomrule
  \end{tabular}
  \caption{\textbf{Results Comparison.} 
  Our solution achieves 76.347 IoU on the private test set, ranking first in the challenge. }
  \label{tab:res}
  \vspace{-1em}
\end{table}

\vspace{-0.5em}
\section{Conclusion}
\label{sec:conc}

This report summarizes our solution for the WSDM2023 Toloka VQA Challenge, which includes three core components: Uni-Perceiver, ViT-Adapter, and DINO. 
Our solution indicates that in addition to uni-modal tasks, ViT-Adapter is also an effective paradigm for adapting the unified perception model to vision-language downstream tasks.
Extensive experiments show the effectiveness of our solution, and it finally won the championship in the challenge.

{\small
\bibliographystyle{ieee_fullname}
\bibliography{egbib}
}

\end{document}